\documentclass[final]{cvpr}

\usepackage{times}
\usepackage{epsfig}
\usepackage{graphicx}
\usepackage{amsmath}
\usepackage{amssymb}

\usepackage{xspace}
\usepackage{marginnote}
\usepackage{mathtools}
\usepackage{microtype}
\usepackage{booktabs} 
\usepackage{multirow}
\usepackage{etoolbox,siunitx}
\usepackage{enumitem}
\usepackage{calc}

\usepackage[pagebackref=true,breaklinks=true,colorlinks,bookmarks=false]{hyperref}

\def\cvprPaperID{6964} 
\def\confYear{CVPR 2021}

\newcommand{\boldentry}[1]{%
  \multicolumn{1}{S[table-format=2.2,
                    mode=text,
                    text-rm=\fontseries{b}\selectfont
                   ]}{#1}}

\begin{document}

\title{Generative Layout Modeling using Constraint Graphs}


%
\author{
Wamiq Para$^1$ \hspace{10pt} Paul Guerrero$^2$ \hspace{10pt} Tom Kelly$^3$ \hspace{10pt} Leonidas Guibas$^4$ \hspace{10pt} Peter Wonka$^1$\\
$^1$KAUST \hspace{10pt} $^2$Adobe Research \hspace{10pt} $^3$ University of Leeds \hspace{10pt} $^4$ Stanford University\\
{\tt\scriptsize \{wamiq.para, peter.wonka\}@kaust.edu.sa \hspace{1pt} guerrero@adobe.com \hspace{1pt} twakelly@gmail.com \hspace{1pt} guibas@cs.stanford.edu}
}

\maketitle

\begin{abstract}
We propose a new generative model for layout generation. We generate layouts in three steps. First, we generate the layout elements as nodes in a layout graph. Second, we compute constraints between layout elements as edges in the layout graph. Third, we solve for the final layout using constrained optimization. For the first two steps, we build on recent transformer architectures. The layout optimization implements the constraints efficiently.
We show three practical contributions compared to the state of the art:
our work requires no user input, produces higher quality layouts, and enables many novel capabilities for conditional layout generation.



\end{abstract}

\section{Introduction}
We study the problem of topologically and spatially consistent layout generation. This problem arises in image layout synthesis, floor plan synthesis, furniture layout generation, street layout planning, and part-based object creation, to name a few.
Generated content must meet stringent criteria both globally, in terms of its overall \textit{topological} structure, as well as locally, in terms of its \textit{spatial} detail.
While our work applies to layouts in general, we focus our discussion on two types of layouts: floorplans and furniture layouts.
%


When assessing layouts, we must consider the global structure which is largely topological in nature, such as connectivity between individual elements or inter-element hop distance.
We are also concerned with spatial detail, such as 
the geometric realization of the elements and their relative positioning, both local and non-local. Realism of such generated content is often assessed by comparing distributions of their properties, both topological and spatial, against those from real-world statistics.



\begin{figure}[t]
    \centering
    \includegraphics[width=\columnwidth]{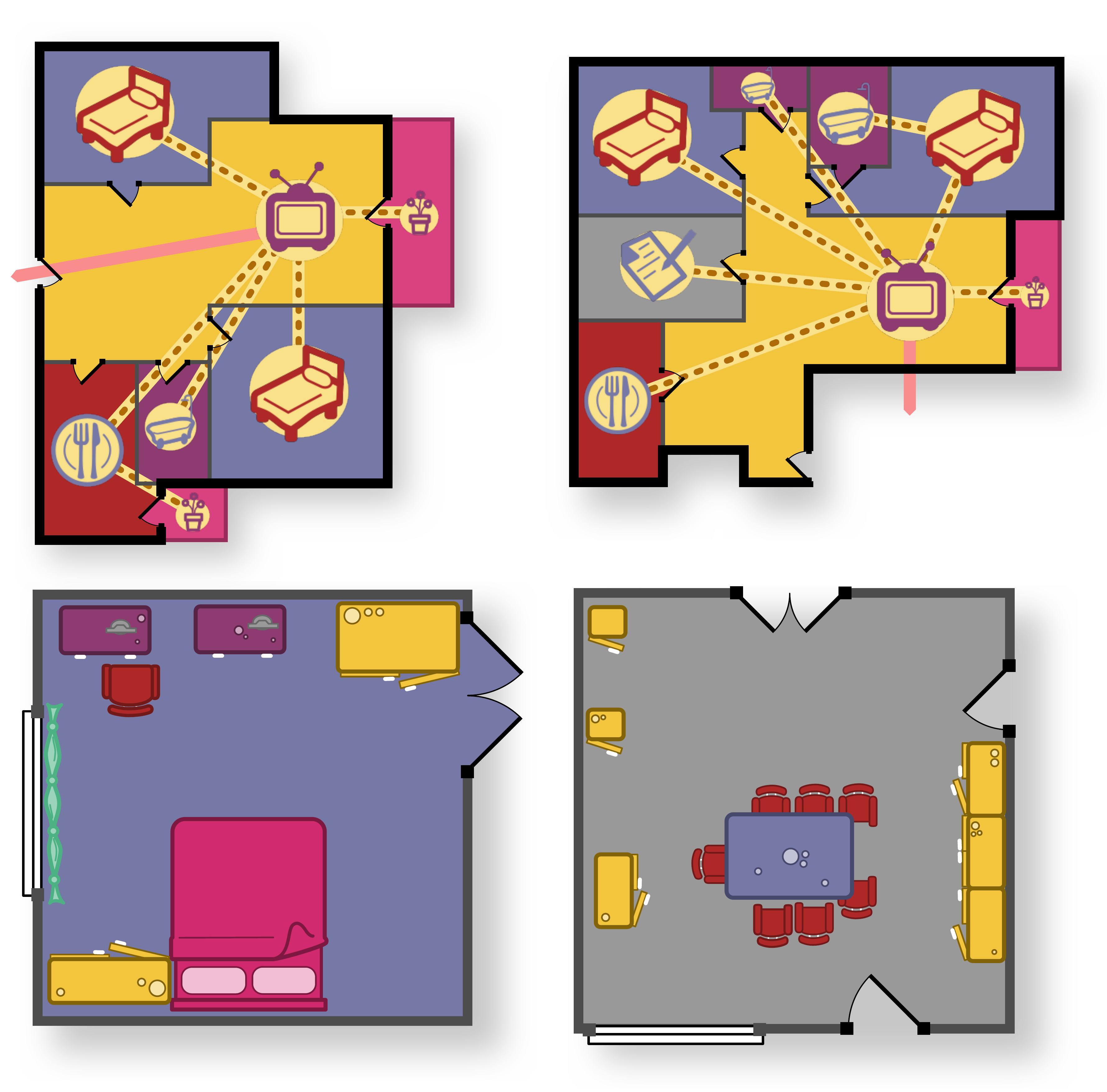}
    \caption{We present a method for layout generation. Our approach can generate multiple types of layouts, such as the floor plans in the top row, where rooms are colored by type, and furniture layouts in the bottom row, where furniture pieces are colored by type. Layouts are represented as graphs, where nodes correspond to layout elements and edges to relationships between elements. In the top row, nodes represent rooms (illustrated with room-type icons), and edges relate rooms connected by doors (dotted lines). Unlike previous methods, our method does not require any input guidance and generates higher-quality layouts.}
    \label{fig:teaser}
\end{figure}

Techniques for synthesizing realistic content have made rapid progress in recent years due to the emergence of generative adversarial networks (GANs)~\cite{Goodfellow2014GANs,Zhu2017CycleGAN,styleGan18,Wang:2018:AAAI:graphGAN}, variational autoencoders (VAEs)~\cite{kingma2013auto, vahdat2020NVAE}, flow models~\cite{rezende2015variational, yang2019pointflow, vahdat2020NVAE}, and autoregressive models~\cite{Chen2020GPTImage}. However, satisfying \textit{both} topological and spatial properties still remains an open challenge. 

Recently, three  papers targeting this challenging problem in the floor plan setting were published~\cite{Wu_DeepLayout_2019, Graph2Plan20, housegan2020}. While these papers often produce good looking floor plans, they require several simplifications to to tackle this difficult problem:
1) RPLAN~\cite{Wu_DeepLayout_2019} and Graph2Plan~\cite{Graph2Plan20} require the outline of the floorplan to be given.
2) HouseGAN~\cite{housegan2020} does not generate the connectivity between rooms that would be given by doors, and RPLAN places doors using a manually defined heuristic that is not learned from data.
3) HouseGAN and Graph2Plan require the number of rooms, the room types and their topology to be given as input in the form of an adjacency graph.
4) All three methods require a heuristic post-process that is essential to make the floorplan look more realistic, but that is not learned from data. In addition, there is still a lot of room to improve the quality and realism of the results.

In this paper, we would like to explore two ideas to improve upon this exciting initial work.
First, after extensive experiments with many variations of graph-based GANs and VAEs, we found that these architectures are not well suited
to tackle the problem. It is our conjecture that these methods struggle with the discrete nature of graphs and layouts.
We therefore propose
an auto-regressive model using attention and self-attention layers. Such an architecture inherently handles discrete data and gives superior performance to current state of the art models. While transformer-based auto-regressive models~\cite{vaswani2017transformers} just started to compete with GANs built on CNNs in image generation~\cite{parmar2018image, chen2020generative} on the ImageNet~\cite{imagenet_cvpr09} dataset, the gap between these two competing approaches for \emph{layout generation} is significant.

Second, we explore the idea of generative modeling using constraint generation. We propose to model layouts with autoregressive models that generate constraint graphs: individual shapes are nodes and edges between nodes specify constraints.
Our auto-regressive model first generates initial nodes, that are subsequently optimized to satisfy constraint edges generated by a second auto-regressive model.
These models can be conditioned on additional constraints provided by the user. This enables various forms of conditional generation and user interaction, from satisfying constraints provided by the user, to a fully generative model that generates constraints from scratch without user interaction.
For example, a user can optionally specify a floorplan boundary, or a set of rooms.

We demonstrate our approach in the context of floor plan generation by creating apartment-level room layouts and furniture layouts for each of the generated rooms (see Figure~\ref{fig:teaser}). Our evaluation will show that our generative model allows layout creation that matches both global and local statistics of real-world data much better than competing work. 



In summary, we introduce two main contributions: 1) A transformer-based architecture for generative modeling of layouts that produces higher quality layouts than previous work. 2) The idea of a generative model that generates constraint graphs and solves for the spatial shape attributes via optimization, rather than outputting shapes directly.

\section{Related Work}

We will discuss image-based generative models, graph-based generative models, and finally models specialized to layout generation.


\subsection{Image-based Generation}
A straight-forward approach to generate a layout is to represent it as an image and use traditional generative models for image synthesis.
The most promising approach are generative adversarial networks~(GANs)~\cite{Goodfellow2014GANs,Karras2017ProgressiveGrowing,Han18SAG,DBLP:journals/corr/abs-1809-11096,2018arXiv181204948K,SGv2,Karras2020ada}.
Image-to-image translation GANs~\cite{Isola2016Pix2Pix, Zhu2017CycleGAN, NIPS2017_6650, Huang:2018:MUNIT, zhu2019sean,richardson2020encoding} could also be useful for layout generation, e.g., as demonstrated in this project~\cite{chaillou2020archigan}.
Alternatively, modern varitional autoencoder, such as NVAE~\cite{vahdat2020NVAE} or VQ-VAE2~\cite{DBLP:conf/nips/RazaviOV19} are also viable options.
Autoregressive models, e.g.~\cite{Chen2020GPTImage}, also showed great results on larger datasets recently.
When experimenting with image-based GANs, we noticed that they fail to respect the relationships between elements and that they cannot preserve certain shapes (e.g. axis-aligned polygons, sharp corners).



\subsection{Graph-based Generation}
In order to capture relationships between elements, various graph-based generative models have been proposed~\cite{Wang:2018:AAAI:graphGAN, Li:2018:LDGG, Simonovsky:2018:GraphVAE, You:2018:GraphRNN, GraphFlows2019, GraphAttention2019, GraphEmbedding2020}.
However, purely graph-based approaches only generate the graph topology, but are missing the spatial embedding. The specialized layout generation algorithms described next often try to combine graph-based and spatial approaches.

\begin{figure*}[t]
    \centering
    \includegraphics[width=\textwidth]{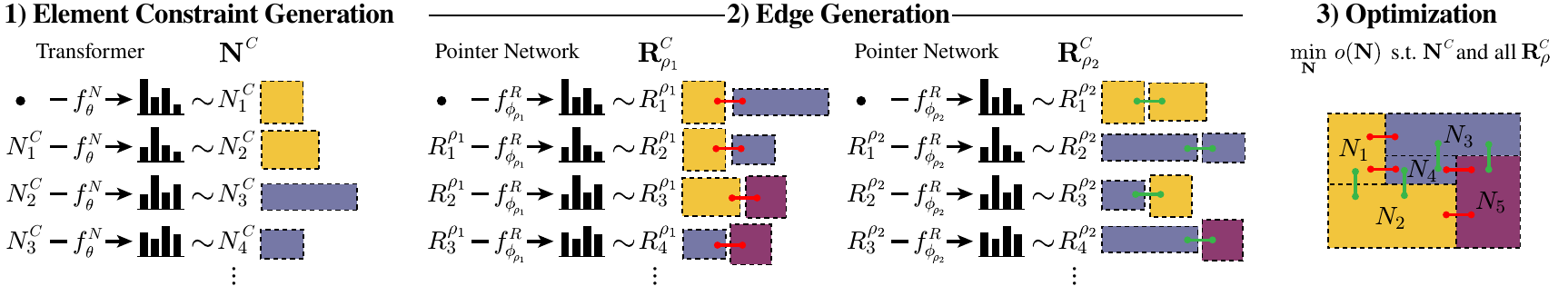}
    \caption{Overview of our layout generation approach. We generate constraints on the parameters of layout elements with a Transformer~\cite{vaswani2017transformers}, and constraints on multiple types of relationships between elements using Pointer Networks~\cite{vinyals2015pointer}. Both element and relationship constraints are used in an optimization to create the final layout.}
    \label{fig:overview}
\end{figure*}

\subsection{Specialized Layout Generation}

Before the rise of deep learning, specialized layout generation approaches have been investigated in numerous domains, including street networks~\cite{Yang:2013:UPL,Peng:2016:CND}, parcels~\cite{bymw_goodLayout_sigg13,Vanegas:2012:PGO}, floor plans~\cite{Wu:2018:MLD}, game levels~\cite{Yeh:2012:SOW}, furniture placements~\cite{Yu:2011:MakeItHome}, furniture and object arrangements~\cite{Fisher:2012:ESO}, and shelves~\cite{Majerowicz:2014:FYS}.
Different approaches have been proposed for layout generation, such as rule-based modeling~\cite{Prusinkiewicz:1990:ABP,Mueller:2006:PMO}, stochastic search ~\cite{Merrell:2011:IFL,Yu:2011:MHA, Yeh:2012:SOW}, or integer programming~\cite{Peng:2014:CLW,Peng:2016:CND,Wu:2018:MLD}, or graphical models~\cite{Merrell:2010:CGR,Fan:2016:APM,Chaudhuri:2011:PRF,Kalogerakis:2012:APM, Fisher:2012:ESO, Yeh:2013:STP}.

In recent years, most of the focus has shifted to applying deep learning to layout generation.
A popular and effective technique
places elements one-by-one, ~\cite{Wang:2018:DCP, LayoutVAE:ICCV:2019, NTG:ICCV:2019}, while a different approach first generates a layout graph and then instantiates elements according to the graph~\cite{IGS:CVPR:2018,PlanIT:2019:Wang,SOA:ICCV:2019}. Both of these approaches are problematic in layouts such as floor plans, that have many constraints between elements, such as zero-gap adjacency and door connectivity.
In such a settings it is non-trivial to a) train a network to generate constraints that admit a solution, and b) find elements that satisfy the constraints in a single forward pass. 
Recently proposed methods~\cite{Wu_DeepLayout_2019, Graph2Plan20, housegan2020} circumvent these problems by requiring manual guidance as input, or by requiring manual post-processing. Due to these requirements, these methods are not fully generative.
%
%
%
%
Recently, Xu et al. introduced PolyGen~\cite{nash2020polygen}, a method to generate graphs of vertices that form meshes with impressive detail and accuracy.
We base our method on a similar architecture, but generate layout constraints instead of directly generating the final layout. Layout elements are then found in an optimization step based on the generated constraints. This gives us layouts where elements accurately satisfy the constraints.

\section{Method}
%
%

We present a generative model for layouts that can optionally be conditioned on constraints given by the user. Figure~\ref{fig:overview} illustrates our approach. Layouts are represented as graphs, where nodes correspond to discrete elements of the layout, and edges represent relationships between the elements.
We distinguish two types of edges: \emph{Constraining edges} describe desirable relationships between element parameters, such as an adjacency between a bedroom and a bathroom in a floor plan, and can be used to constrain these parameters. \emph{Descriptive edges} represent additional properties of the layout that are not given by the elements, but can be useful for down-stream tasks, such as the presence of a door between two rooms of a floor plan where the elements consist of rooms.
%
Section~\ref{sec:layout_rep} describes this layout representation.



A generative model can be trained to generate both layout elements and edges. However, generated elements and generated constraining edges are not guaranteed to match. For example, two elements that are connected by an adjacency edge can often be separated by a gap, or can have overlaps. As the number of constraining edges increases, the problem of generating a compatible set of edges and elements becomes increasingly difficult to solve in a forward pass of the generative model. This has been a major limitation in previous work.

We introduce two contributions over previous layout generation methods. 
First, we show that a two-step autoregressive approach inspired by PolyGen~\cite{nash2020polygen} that first generates elements and then edges is particularly suitable for layout generation and performs significantly better than current methods. We describe this approach in Sections~\ref{sec:element_constraint_model} and~\ref{sec:edge_model}.

Second, we treat element parameters and constraining edges that were generated in the first two steps as \emph{constraints} and optimize element parameters to satisfy the generated constraints in a subsequent optimization step.
In floor plans, for example, we generate constraints on the maximum and minimum widths and heights of room areas and on their adjacency, and then solve for their locations, widths and heights in the optimization step.
This minimizes any discrepancies between constraining edges and element parameters.
%
%
%
%
We describe the optimization in Section~\ref{sec:optimization}.
In Section~\ref{sec:user_constraints}, we describe how to condition on user-provided constraints.

\subsection{Layout Representation}
\label{sec:layout_rep}

We represent layouts as a graph $\mathcal{L}=(\mathbf{N}, \mathbf{R})$, where nodes correspond to layout elements $\mathbf{N}$ and edges to their relationships $\mathbf{R}$. Each layout element $N \in \mathbf{N}$ has a fixed set of domain-specific parameters. Relationship edges $R \in \mathbf{R}$ are chosen from a fixed set of edge types $\rho$ and describe the presence of that edge between two elements $R = (N_i, N_j, \rho)$.
Edges come in two groups, based on their types: constraining edges $\mathbf{R}^C$ that provide constraints for the optimization step, and descriptive edges $\mathbf{R}^D$ that provide additional information about the layout.
%
We consider two main layout domains in our experiments: floor plans and furniture layouts, but will only focus on floor plans here. Furniture layouts are described in the supplementary material.

\begin{figure}
    \centering
    \includegraphics[width=\columnwidth]{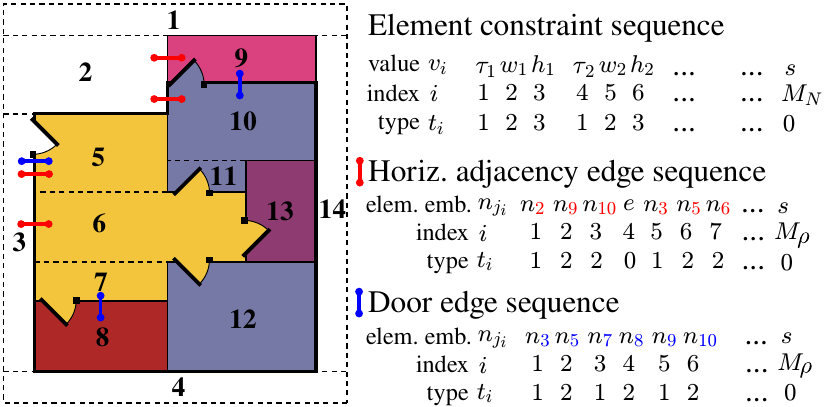}
    \caption{Example floor plan layout and its sequence encoding. Rooms are represented by rectangles, which are numbered and colored by room type for illustration (white for the exterior). Edges either constrain rectangles, like the red adjacency edges, or add information to the layout, like the blue door edges.
    Both are encoded into sequences that can be ingested by our autoregressive sequence-to-sequence models.}
    \label{fig:layout_examples}
\end{figure}

In floor plans, each layout element is a rectangular region of a room $N=(\tau, x,y,w,h)$, parameterized by the type of room $\tau$, the lower-left corner of the rectangular region $(x,y)$, and the width and height $(w,h)$ of the region. Two types of edges in $\mathbf{R}^C$ define horizontal and vertical adjacency constraints between elements, while two types of edges in $\mathbf{R}^D$, define the presence of a wall between two adjacent elements, and the presence of a door between two adjacent elements. Multiple elements of the same type that are adjacent and not separated by a wall form a room. The set of all elements fully cover the floor plan. An example is shown in Figure~\ref{fig:layout_examples}, left.
%
More details on both representations, including a full list of all element types, are given in the supplementary material.



\subsection{Element Constraint Model}
\label{sec:element_constraint_model}

An element constraint $N^C$ is defined as a tuple of target values for one or more of the parameters of element $N$. In the optimization, we will use these values as soft constraints for the corresponding parameters. We create one set of constraints for each element $N$ of the layout. In floor plans, for example, we create constraints $N^C = (\tau, w, h)$ for the type, width, and height of each element. All continuous values are treated as range constraints, i.e. the actual values may be within the range $\pm \epsilon v^C$ of the constraint value $v^C$ (we set $\epsilon=0.1$ in our experiments).
We use a transformer-based~\cite{vaswani2017transformers} autoregressive sequence-to-sequence model to generate these element constraints.

\paragraph{Sequence encoding}
The goal of our element constraint model is to learn a distribution over constraint sequences. To flatten our list of element constraints, we order them from left to right first (small to large $x$) and top to bottom (small to large $y$) for elements with the same $x$ coordinate. The ordered constraint tuples are concatenated to get a sequence of constraint values $S_E = (v_i)_{i=1}^{k M_N}$, where $M_N$ is the number of elements in the layout and $k$ the number of properties per element.
Following PolyGen \cite{nash2020polygen}
we use two additional inputs per token in the sequence:
the sequence index $i$ and the type $t_i$ of each value.
Type $t_i$ is the index of a constraint value inside its constraint tuple and indicates the type of the value (x-location, height, angle, etc.). Finally, we add a special stopping token $s$ as last element of the sequence to indicate the end of the sequence.

\paragraph{Autoregressive Model}
Our element constraint model $f_{\theta}^N$ models the probability of a sequence by its factorization into step-wise conditional probabilities:
\begin{equation}
    p(S_N; \theta) = \prod_{i=1}^{k M_N} p(v_i | v_{<i}; \theta),
\end{equation}
where $\theta$ are the parameters of the model.
Given a partial sequence $v_{<i}$, the model predicts a distribution over values for the next token in the sequence $p(v_i| v_{<i}; \theta) = f_{\theta}^N(v_{<i}, (1 \dots i-1), t_{<i})$, that we can sample to obtain $v_i$. We implement $f_\theta$ with a small version of GPT-2~\cite{radford2019language} that has roughly 10 million parameters. For architecture details, please refer to Section \ref{sec:architecture} and the supplementary material.

\paragraph{Coordinate Quantization}
We apply 6-bit quantization for all coordinate values except $\alpha$, which we quantize to 5 bits. We learn a categorical distribution over the discrete constraint values in each step of the model. Nash et al.~\cite{nash2020polygen} have shown that this improves model performance, since it facilitates learning distributions with complex shapes over the constraint values.



\subsection{Edge Model}
\label{sec:edge_model}
We generate relationship edges $R$ between elements that either constrain element parameters or add additional information to the layout. The constraining edges $\mathbf{R}^C$ will be used as constraints during the optimization step, while descriptive edges $\mathbf{R}^D$ add information to the layout and may be needed in down-stream tasks. In floor plans, for example, door and wall edges define walls and doors. We use an autoregressive sequence-to-sequence architecture based on PointerNetworks~\cite{vinyals2015pointer} to generate edges. We train one model for each of the edge types described in Section~\ref{sec:layout_rep}, each models the distribution for one type of edge. All models have the same architecture, but do not share weights.

\paragraph{Sequence Encoding} To flatten the list of edges $R = (N_i, N_j, \rho)$ of any given type $\rho$, we first sort them by the index of the first element $i$, then by the index of the second element $j$.
We then concatenate the constraints $N^C_i$, $N^C_j$ corresponding to the elements $N_i, N_j$ in each edge to get a sequence of element constraints. We use a learned embedding $n_j^{\rho} = g_{\phi_{\rho}}(N^C_j)$, giving us a sequence of element embeddings $S_{\rho} = (n_{j_i}^{\rho})_{i=1}^{2 M_{\rho}}$, where $M_{\rho}$ is the number of edges of a given type $\rho$.
%
Two additional inputs are added for each token: the index $i$ and the type $t_i$, indicating if a token corresponds to the source or target element of the edge. The last token in the sequence is the stopping token $s$.

Due to our ordering, groups of edges that share the same source element $N_i$, are adjacent in the list.
For types of edges where these groups are large, that is, where many edges share the same source element, we can shorten the sequence by including the constraint of a source element only once at the start of the group, and then listing only the constraints of the target elements $N_j$ that are connected to this source element. The end of a group is indicated by a special token $e$. We use this shortened sequence style for the adjacency edges of floor plans.

\paragraph{Autoregressive Model}
Similar to the element constraint model, the probability of an edge sequence $S_{\rho}$ is modeled by a factorization into step-wise conditional probabilities. Unlike the element constraint model, however, the edge model $f_{\phi_{\rho}}^R$ outputs a pointer embedding~\cite{vinyals2015pointer}:
\begin{equation}
    q_i^{\rho} = f_{\phi_{\rho}}^R(n_{j_{<i}}^{\rho}, (1 \dots i-1), t_{<i}).
\end{equation}
We compare this pointer embedding to all element embeddings using a dot-product to get a probability distribution over elements:
\begin{equation}
    p(n_{j_i}^{\rho}{=}n_k^{\rho} | n_{j_{<i}}^{\rho}; \phi_{\rho}) = \mathrm{softmax}_k\left((q_i^{\rho})^T n_k\right)
\end{equation}
that we can sample to get the index of the next element constraint in the sequence.

\subsection{Optimizing Layouts}
\label{sec:optimization}

We formulate a Linear Programming problem~\cite{boyd2004convex} that regularizes the layout while satisfying all generated constraints:
\begin{equation}
\begin{aligned}
    \min_{\mathbf{N}} \quad & o(\mathbf{N}) \\
    \textrm{s.t.} \quad & \mathbf{N}^C\ \text{are satisfied and}\\
                        & \mathbf{R}^C\ \text{are satisfied},
\end{aligned}
\end{equation}
where $o(\mathbf{N})$ is a regularization term. In floor plans, for example, we minimize the perimeter of the floor plan $o(\mathbf{N}) = W + H$, where $W$ and $H$ are the width and height of the floor plan's bounding box. This effectively minimizes the size of the layout, while keeping the optimization problem linear. This regularization encourages compactness and a bounded layout size, resulting in layouts without unnecessary gaps and holes. The definition of the constraints depend on the type of layout.


In floor plans, the $x,y,w,h$ parameters of each element are bounded between their maximum and minimum values; we use $[0, 64]$ as bounds in our experiments.
Each element constraint $N^C$ adds constraints of the form $v^C (1-\epsilon) \leq v \leq v^C (1+\epsilon)$, for each value $v^C$ in the element constraint $N^C$ and corresponding value $v$ in the element $N$. In our experiments, we set $\epsilon = 0.1$.
Horizontal adjacency edges $R = (N_i, N_j, \rho)$ add constraints of the form $x_i + w_i = x_j$, and analogously for vertical adjacency edges.

The layout width $W$ is computed by first topologically sorting the elements in the subgraphs formed by horizontal adjacency edges, and then defining $W \coloneqq x_m + w_m$ for the last (right-most) element $N_m$ in the topological sort. $H$ is computed analogously. Note that we do not define $W \coloneqq \max_i x_i + w_i$ to avoid the additional constraints needed to optimize over the maximum of a set. A detailed list of constraints for furniture layouts is given in the supplementary material. The challenge of designing the optimization is to keep the optimization fast and simple
and to make it work in conjunction with the neural networks.

\subsection{User-provided Constraints}
\label{sec:user_constraints}

We can condition our models on any user-provided element constraints.
We add an encoder
to both the element constraint model and the edge model,
following the encoder/decoder architecture described in~\cite{vaswani2017transformers}. The
encoder
takes as input a flattened sequence
of user-provided constraints, enabling cross-attention from the sequence
that is currently being generated to the list of user constraints.
Note that the user-provided constraints do not have to represent the same quantities as the output sequence. In floor plans, for example, we can condition both the element constraint model and the edge model on a list of room types, room areas and/or a floor plan boundary.




\subsection{Network Architecture}
\label{sec:architecture}
Our models use the Transformer \cite{vaswani2017transformers} as a building block. Our Element Constraint Model and the Edge model are very similar to the Vertex and Face models from PolyGen \cite{nash2020polygen} in organization. The building block for the Transformers themselves is based on the GPT-2 model, specifically, we use the GELU activation \cite{hendrycks2016gelu}, Layer Norm \cite{layernorm19} and Dropout. For a complete description, please refer to the supplementary.

The model for element constraint generation consists of 12 Transformers blocks. Our sequence lengths depend on the particular dataset used, and are listed in the supplementary.
The edge generation model is a Pointer Network with two-parts: 1. An encoder which generates embeddings, and can attend to all elements in the sequence of element constraints and 2. A decoder which generates pointers, and can attend to elements in an autoregressive fashion. In our experiments, we use an encoder with 16 layers and a decoder with 12 layers. 
We use 384 dimensional embeddings in all our models.

\begin{figure}[t]
    \centering
    \includegraphics[height=1.7in]{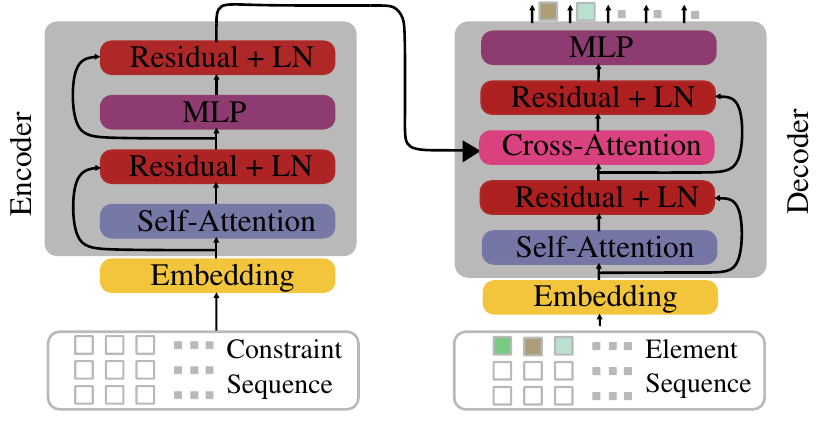}
    \caption{The constrained element generation model. Having an unmasked encoder allows our network to attend to all elements of the constraint sequence.
    }
    \label{fig:enc_dec_model}
\end{figure}

Constrained generation 
is performed by a variant of the unconstrained models. Concretely, we add a constraint encoder to both the element constraint model and the edge models resulting in an encoder-decoder architecture. In the edge models, we concretely change the encoder of the Pointer Network to an encoder-decoder architecture.
(Figure \ref{fig:enc_dec_model}).
The constraint encoder
is a stack of Transformer blocks allowed to attend all elements of the constraint sequence. The decoder
is another stack of blocks allowed to attend to all tokens in the constraint sequence.
We use 8 layers for constraint encoder in the element model and 3 layers in the edge-model.

\paragraph{Training Setup}
\label{sec:traning_setup}
We implemented our models in Pytorch\cite{paszke2017automatic}. Our models and sequences are small enough so we train on a single NVIDIA-V100 GPU with 32 GB memory. We use the Adam \cite{DBLP:journals/corr/KingmaB14} optimizer, with a constant learning-rate of $10^{-4}$, and linear warmup for 500 iterations. The element generation model is trained for 40 epochs, while the other models are trained for 80 epochs. It takes approximately 6 hours to train for our largest model for constrained generation.

\paragraph{Inference} The inference time depends on the type of sequence being sampled. Our large sequences have about 250 tokens. For this sequence length, generating a batch of 100 element constraint sequences  takes about 10s. Given the element constraint sequence, all types of edges can be sampled in parallel. Edge models are larger and need about 60s for a batch of 100 sequences.


\section{Results}
\label{sec:results}
We evaluate free generation of layouts, generation constrained by a given boundary, and generation constrained by additional user-provided constraints. We will focus on floor plans in this section. Furniture layouts are evaluated in the supplementary material.
\paragraph{Datasets}
We train and evaluate on two floor plan datasets.
The RPLAN dataset~\cite{Wu_DeepLayout_2019} contains 80k floor plans of apartments or residential buildings in an Asian real estate market between 60$\text{m}^2$ to 120$\text{m}^2$.
%
The LIFULL dataset~\cite{lifull} contains 61k floor plans of apartments from the Japanese housing market. The apartments in this dataset tend to be more compact.
The original dataset is given as heterogeneous images, but a subset was parsed by Liu et al.~\cite{Liu:2017:RasterToVector} into a vector format.
In both datasets we use 1k layouts for each of testing and validation, and the remainder for training.
\paragraph{Baselines}
StyleGAN~\cite{styleGan18} generates a purely image-based representation of a layout. We render the layout into an image to obtain a training set, including doors and walls (see the supplementary material), and parse the generated images to obtain layouts. Graph2Plan~\cite{Graph2Plan20} generates a floor plan given its boundary and a layout graph that describes rough room locations, types, and adjacencies. Door connectivity is generated heuristically. RPLAN~\cite{Wu_DeepLayout_2019} generates a floor plan given its boundary, with a heuristically-generated door connectivity. All baselines are re-trained on each dataset.
\begin{table}[t]
\caption{Free generation of layouts. We compare the FID and layout statistics on two datasets to the state-of-the-art. Note that Graph2Plan uses a ground-truth layout graph as input, and both RPLAN and Graph2Plan use the ground truth boundary as input. We evaluate both free generation with our method and conditional generation. Our method improves upon the baselines with less input guidance.}
\centering 
\footnotesize
\setlength{\tabcolsep}{5pt}
\begin{tabular}{@{}llS[table-format=2.2]S[table-format=2.2]S[table-format=2.2]S[table-format=2.2]S[table-format=2.2]c@{}} 
\toprule
dataset & method & {FID} & $\hat{s}_{t}$ & $\hat{s}_{r}$ & $\hat{s}_{a}$ & $\hat{\mathbf{s}}_\textbf{avg}$ \\
\midrule \midrule
\multirow{4}{*}{RPLAN}
& StyleGAN. & 25.29 & 46.74 & 4.41 & 7.85 & 19.67 & \\
& Graph2Plan & 29.26 & 0.83 & 5.63 & 18.93 & 8.46 & \\
& RPLAN & \boldentry{21.29} & 5.38 & 1.53 & 4.38 & 3.76 & \\
& ours free & 21.47 & 1.00 & 1.00 & \boldentry{1.00} & \boldentry{1.00} & \\
& ours cond. & 27.27 & \boldentry{0.81} & \boldentry{0.94} & 1.34 & 1.03 & \\
\midrule
\multirow{4}{*}{LIFULL}
& StyleGAN & 28.06 & 44.54 & 2.32 & 1.96 & 16.27 & \\
& Graph2Plan & 29.50 & 9.21 & 0.94 & 1.37 & 3.84 & \\
& RPLAN & 32.98 & 40.54 & 2.02 & 4.10 & 15.55 & \\
& ours free & \boldentry{26.15} & 1.00 & 1.00 & 1.00 & \boldentry{1.00} & \\
& ours cond. & 31.94 & 5.70 & \boldentry{0.71} & \boldentry{0.50} & 2.30 & \\
\bottomrule
\end{tabular}

\label{tab:free_generation}
\end{table}

\begin{figure*}[t]
    \centering
    \includegraphics[width=\textwidth]{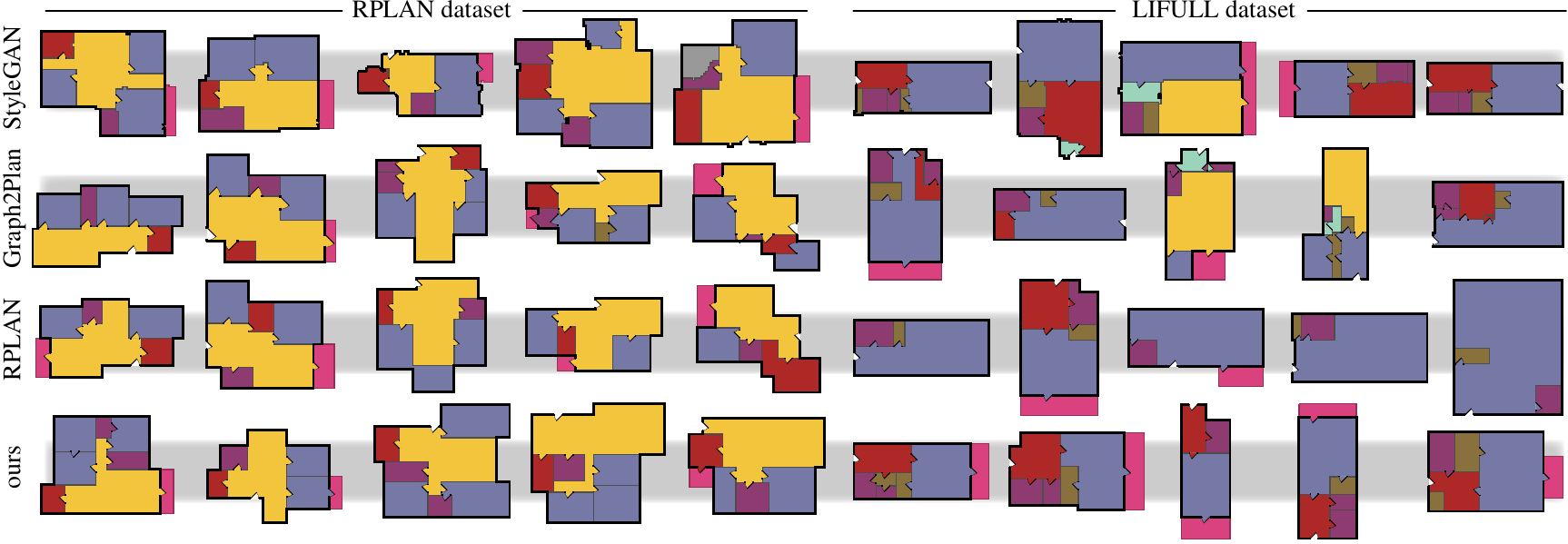}
    \caption{Free generation of floor plans. We compare our method to three baselines. Rooms are colored by room type, with an overlaid door connectivity graph. Nodes on the graph have icons based on the room type. Our three-step approach improves upon the room layout and connectivity compared to previous approaches, while requiring less guidance as input.}
    \label{fig:free_generation}
\end{figure*}
\paragraph{Metrics} We compare generated layouts to ground truth layouts using two metrics: The \emph{Fr\'echet Inception Distance} (FID)~\cite{heusel2017gans} computed on rendered layouts, and a metric based on a set of \emph{layout statistics} that measure layout properties that the image-based FID is less suitable for. Layout statistics are grouped into topological statistics $S_t$ such as the average graph distance in the layout graph between any two element types,
element shape statistics $S_r$ such as the aspect ratio or area, and alignment statistics $S_a$ such as the gap between adjacent elements, or their boundary alignment. We believe that our proposed statistics are more useful to evaluate layouts than FID. FID is more suitable to evaluate generative models trained on natural images, but we show the FID metric for completeness as it is more widely used.

{\em Topological statistics} $S_t$ are specialized to measure the topology of a layout graph~\cite{bookStreet,Tarabishy2019DeepLS}:
\begin{itemize}[itemsep=-3pt]
    \small
    \item[$s_t^r$:] the average number of elements of a given type in a layout.
    \item[$s_t^h$:] a histogram over the number of elements of a given type in a layout.
    \item[$s_t^t$:] the number of connections between elements of type $a$ and elements of type $b$ in a layout.
    \item[$s_t^d$:] the average graph distance between elements of type $a$ and elements of type $b$ in a layout.
    \item[$s_t^e$:] a histogram of the graph distance from an element of type $a$ to the exterior.
    \item[$s_t^c$:] a histogram of the degree of an element of type $a$, i.e. how many connections the element has to other elements.
    \item[$s_t^u$:] The number of inaccessible elements of type $a$ in a layout.
\end{itemize}
{\em Element shape statistics} $S_r$ measure simple properties of the element bounding boxes:
\begin{itemize}[itemsep=-3pt]
    \small
    \item[$s_r^c$:] a histogram of location distributions for each element type.
    \item[$s_r^a$:] a histogram of area distributions for each element type.
    \item[$s_r^s$:] a histogram of aspect ratio distributions for each element type.
\end{itemize}
{\em Alignment statistics} $S_a$ measure alignment between all pairs of elements:
\begin{itemize}[itemsep=-3pt]
    \small
    \item[$s_a^c$:] a histogram of the distances between element centers, separately in x and y direction.
    \item[$s_a^g$:] a histogram of the gap size distribution between element bounding boxes (negatives values for overlaps).
    \item[$s_a^a$:] a histogram of the distances between element centers along the best-aligned (x or y) axis.
    \item[$s_a^s$:] a histogram of the distances between the best-aligned sides of the element bounding boxes.
\end{itemize}
The same alignment statistics are also computed between pairs of elements that are connected by descriptive edges.

We average each statistic over all layouts in a dataset and compare the resulting averages $\overline{s}$ to the statistics of the test set. We use the Earth Mover's distance~\cite{Rubner:1998:MDA} to compare histograms:
\begin{equation}
\hat{s}_{*} = \frac{1}{|S_*|}\sum_{s \in S_*}\frac{\text{EMD}(\overline{s}, \overline{s}^\text{gt})}{\text{EMD}(\overline{s}^\text{ours}, \overline{s}^\text{gt})},
\end{equation}
where $\overline{s}^\text{ours}$ and $\overline{s}^\text{gt}$ are the average statistics of our and ground truth distributions, and $*$ can be $t$, $r$ or $a$. The average over all $\hat{s}_{*}$ is denoted $\hat{\mathbf{s}}_\textbf{avg}$.
Non-histogram statistics use the L2 distance instead of the EMD.


\paragraph{Free Generation} In a first experiment, we generate floor plans fully automatically, without any user input, by sampling the distribution learned by our constraint model.
A comparison to all baselines is shown in Table~\ref{tab:free_generation} and Figure~\ref{fig:free_generation}. Note that among the baselines, only StyleGAN can generate floor plans without user input, while Graph2Plan and RPLAN need important parts of the ground truth as input. For example, we sample topologies and boundaries from the ground truth and give them to the other methods as input. This gives other methods a significant advantage in this comparison. The FID score correlates most strongly with the adjacency statistics, since adjacencies can be captured by only considering small spatial neighborhoods around corners and walls of a floor plan, but does not capture topology or room shape statics accurately that require considering larger-scale features. Unsurprisingly, StyleGAN performs reasonably well on the FID score and adjacency statistics, but shows a poor performance on topological statistics which are mainly based on larger-scale combinatorial features of the floor plans.
Graph2Plan receives the topology as input giving it a good performance in topological statistics, but it struggles with room alignment. The RPLAN baseline is specialized to the RPLAN dataset, as shown in the large performance gap between RPLAN and LIFULL. In summary, our proposed framework improves significantly on the state-of-the art, in terms of layout topology, element shape, and element alignment, even though RPLAN and Graph2Plan received significant help from ground truth data.

\begin{figure*}[t]
    \centering
    \includegraphics[width=\textwidth]{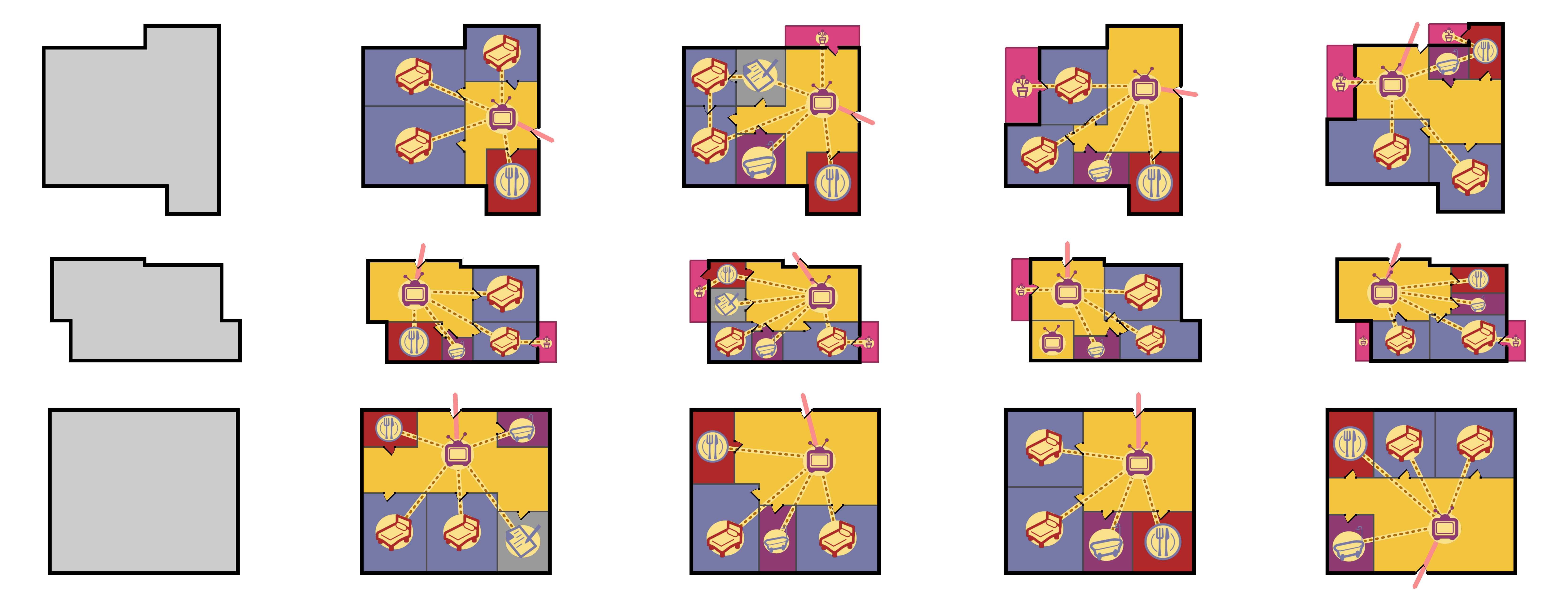}
    \caption{Boundary-constrained generation. Left: input boundary constraint; right: floorplans generated with this constraint.}
    \label{fig:boundary_conditioned_generation}
\end{figure*}
\paragraph{Boundary-constrained Generation} As described in Section~\ref{sec:user_constraints}, we can condition both our element constraint model and our edge model on input constraints provided by the user. Here, we show floor plan generation constrained by an exterior floor plan boundary given by the user. We parse the exterior of the given boundary into a sequence of rectangular elements that we use as input sequence for the encoders of our models. At training time, we use the exterior of ground truth floor plans as input. This trains the models to output sequences of element constraints and edges that are roughly compatible with the given boundary. In the optimization step, we add non-overlap constraints between the generated boxes and the given boundary. Additionally, since the interior boxes are generated in sequence from left to right, we can initialize the first generated box to match the left-most part of the interior area. Figure~\ref{fig:boundary_conditioned_generation} show multiple examples of floor plans that were generated for the boundary given on the left. Quantitative results obtained by conditioning on all boundaries in the test set are provided in the last row for each dataset in Table~\ref{tab:free_generation}. The boundary-constrained floor plans show slightly lower performance in the average layout statistics and FID scores, but still perform much better than RPLAN, which also receives the boundary as input. We can see that our approach gives realistic floor plans that satisfy the given boundary constraint.

\begin{figure}[t]
    \centering
    \includegraphics[width=\columnwidth]{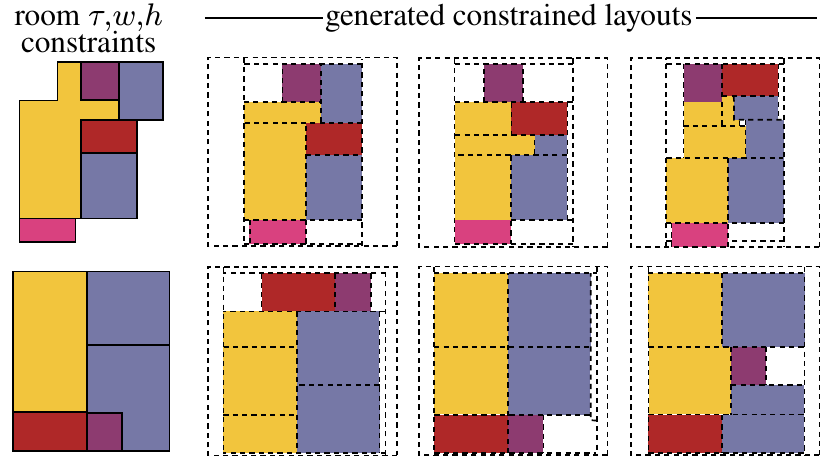}
    \caption{Element-constrained generation. Left: The type, width, and height of the these rooms are used as input constraints. Right: example layouts generated with these constraints. Note that the elements form regions of the same types and approximately the same width and height as the room constraints.}
    \label{fig:room_constrained_generation}
\end{figure}
\paragraph{Element-constrained Generation} Our approach can also handle constraints that are given in a different format than the output. We constrain our model to produce a given set of room types, widths, and heights. Results are shown in Figure~\ref{fig:room_constrained_generation}. Even though these constraints are quite limiting, our model produces a large variety of results, while still approximately satisfying the given constraints. 
\paragraph{Discussion}
Our work also has some limitations. For example, the constraint generation network can generate invalid constraints between elements, e.g. doors between rooms that do not share a wall. We can easily identify and remove these constraints. In addition, some constraints result in optimization problems that are infeasible. We simply ignore such samples. Further, like other methods, our work generates a small percentage of low quality results, however, not nearly as many as other methods, which is reflected in the statistics.

\section{Conclusion}
We proposed a new generative model for layout generation. Our model first generates a layout graph with layout elements as nodes and constraints between layout elements as edges. The final layout is computed by optimization. Our model overcomes many limitations of previous models, mainly the need for significant user input and ad-hoc post-processing steps. Further, our model leads to significantly higher generation quality as evidences by multiple statistics and enables multiple possibilities of conditional layout generation.
In future work, we would like to explore the application of our model to other layout problems, such as image layouts, 3D scene layouts, and component-based object modeling. We also would like to explore if our model can be used to post-process 3D scans of indoor environments. 


{\small
\bibliographystyle{ieee_fullname}
\bibliography{main}

\begin{thebibliography}{10}\itemsep=-1pt

\bibitem{SOA:ICCV:2019}
Oron Ashual and Lior Wolf.
\newblock Specifying object attributes and relations in interactive scene
  generation.
\newblock {\em International Conference on Computer Vision}, 2019.

\bibitem{ba2016layer}
Jimmy~Lei Ba, Jamie~Ryan Kiros, and Geoffrey~E Hinton.
\newblock Layer normalization.
\newblock {\em arXiv preprint arXiv:1607.06450}, 2016.

\bibitem{bymw_goodLayout_sigg13}
Fan Bao, Dong-Ming Yan, Niloy~J. Mitra, and Peter Wonka.
\newblock Generating and exploring good building layouts.
\newblock {\em ACM Transactions on Graphics}, 32(4), 2013.

\bibitem{boyd2004convex}
Stephen Boyd, Stephen~P Boyd, and Lieven Vandenberghe.
\newblock {\em Convex optimization}.
\newblock Cambridge university press, 2004.

\bibitem{DBLP:journals/corr/abs-1809-11096}
Andrew Brock, Jeff Donahue, and Karen Simonyan.
\newblock Large scale {GAN} training for high fidelity natural image synthesis.
\newblock {\em CoRR}, abs/1809.11096, 2018.

\bibitem{chaillou2020archigan}
Stanislas Chaillou.
\newblock Archigan: Artificial intelligence x architecture.
\newblock In {\em Architectural Intelligence}, pages 117--127. Springer, 2020.

\bibitem{Chaudhuri:2011:PRF}
Siddhartha Chaudhuri, Evangelos Kalogerakis, Leonidas Guibas, and Vladlen
  Koltun.
\newblock Probabilistic reasoning for assembly-based {3D} modeling.
\newblock {\em ACM Trans. Graph.}, 30(4):35:1--35:10, July 2011.

\bibitem{chen2020generative}
Mark Chen, Alec Radford, Rewon Child, Jeff Wu, Heewoo Jun, Prafulla Dhariwal,
  David Luan, and Ilya Sutskever.
\newblock Generative pretraining from pixels.
\newblock In {\em Proceedings of the 37th International Conference on Machine
  Learning}, volume~1, 2020.

\bibitem{Chen2020GPTImage}
Mark Chen, Alec Radford, Rewon Child, Jeffrey Wu, Heewoo Jun, David Luan, and
  Ilya Sutskever.
\newblock Generative pretraining from pixels.
\newblock 2020.

\bibitem{NTG:ICCV:2019}
Hang Chu, Daiqing Li, David Acuna, Amlan Kar, Maria Shugrina, Xinkai Wei,
  Ming-Yu Liu, Antonio Torralba, and Sanja Fidler.
\newblock Neural turtle graphics for modeling city road layouts.
\newblock {\em International Conference on Computer Vision}, 2019.

\bibitem{imagenet_cvpr09}
J. Deng, W. Dong, R. Socher, L.-J. Li, K. Li, and L. Fei-Fei.
\newblock {ImageNet: A Large-Scale Hierarchical Image Database}.
\newblock In {\em CVPR09}, 2009.

\bibitem{Fan:2016:APM}
Lubin Fan and Peter Wonka.
\newblock A probabilistic model for exteriors of residential buildings.
\newblock {\em ACM Transactions on Graphics}, 35(5):155, 2016.

\bibitem{Fisher:2012:ESO}
Matthew Fisher, Daniel Ritchie, Manolis Savva, Thomas Funkhouser, and Pat
  Hanrahan.
\newblock Example-based synthesis of 3d object arrangements.
\newblock {\em ACM Trans. Graph.}, 31(6):135:1--135:11, Nov. 2012.

\bibitem{Goodfellow2014GANs}
Ian Goodfellow, Jean Pouget-Abadie, Mehdi Mirza, Bing Xu, David Warde-Farley,
  Sherjil Ozair, Aaron Courville, and Yoshua Bengio.
\newblock Generative adversarial nets.
\newblock In Z. Ghahramani, M. Welling, C. Cortes, N.~D. Lawrence, and K.~Q.
  Weinberger, editors, {\em Advances in Neural Information Processing Systems},
  pages 2672--2680, 2014.

\bibitem{hendrycks2016gelu}
Dan Hendrycks and Kevin Gimpel.
\newblock Gaussian error linear units (gelus).
\newblock {\em arXiv preprint arXiv:1606.08415}, 2016.

\bibitem{heusel2017gans}
Martin Heusel, Hubert Ramsauer, Thomas Unterthiner, Bernhard Nessler, and Sepp
  Hochreiter.
\newblock Gans trained by a two time-scale update rule converge to a local nash
  equilibrium.
\newblock In {\em Advances in Neural Information Processing Systems}, pages
  6626--6637, 2017.

\bibitem{Graph2Plan20}
Ruizhen Hu, Zeyu Huang, Yuhan Tang, Oliver~Van Kaick, Hao Zhang, and Hui Huang.
\newblock Graph2plan: Learning floorplan generation from layout graphs.
\newblock {\em ACM Transactions on Graphics}, 39(4):118:1--118:14, 2020.

\bibitem{Huang:2018:MUNIT}
Xun Huang, Ming{-}Yu Liu, Serge~J. Belongie, and Jan Kautz.
\newblock Multimodal unsupervised image-to-image translation.
\newblock {\em ECCV 2018}, 2018.

\bibitem{Isola2016Pix2Pix}
Phillip Isola, Jun-Yan Zhu, Tinghui Zhou, and Alexei~A Efros.
\newblock Image-to-image translation with conditional adversarial networks.
\newblock {\em arxiv}, 2016.

\bibitem{IGS:CVPR:2018}
Justin Johnson, Agrim Gupta, and Li Fei-Fei.
\newblock Image generation from scene graphs.
\newblock {\em Conference on Computer Vision and Pattern Recognition}, 2018.

\bibitem{LayoutVAE:ICCV:2019}
Akash~Abdu Jyothi, Thibaut Durand, Jiawei He, Leonid Sigal, and Greg Mori.
\newblock Layoutvae: Stochastic scene layout generation from a label set.
\newblock {\em International Conference on Computer Vision}, 2019.

\bibitem{Kalogerakis:2012:APM}
Evangelos Kalogerakis, Siddhartha Chaudhuri, Daphne Koller, and Vladlen Koltun.
\newblock A probabilistic model for component-based shape synthesis.
\newblock {\em ACM Trans. Graph.}, 31(4):55:1--55:11, July 2012.

\bibitem{Karras2017ProgressiveGrowing}
Tero Karras, Timo Aila, Samuli Laine, and Jaakko Lehtinen.
\newblock Progressive growing of gans for improved quality, stability, and
  variation.
\newblock {\em CoRR}, abs/1710.10196, 2017.

\bibitem{Karras2020ada}
Tero Karras, Miika Aittala, Janne Hellsten, Samuli Laine, Jaakko Lehtinen, and
  Timo Aila.
\newblock Training generative adversarial networks with limited data.
\newblock In {\em Proc. NeurIPS}, 2020.

\bibitem{2018arXiv181204948K}
Tero {Karras}, Samuli {Laine}, and Timo {Aila}.
\newblock {A Style-Based Generator Architecture for Generative Adversarial
  Networks}.
\newblock {\em arXiv e-prints}, page arXiv:1812.04948, Dec. 2018.

\bibitem{styleGan18}
Tero Karras, Samuli Laine, and Timo Aila.
\newblock A style-based generator architecture for generative adversarial
  networks.
\newblock {\em CoRR}, 2018.

\bibitem{SGv2}
Tero Karras, Samuli Laine, Miika Aittala, Janne Hellsten, Jaakko Lehtinen, and
  Timo Aila.
\newblock Analyzing and improving the image quality of stylegan.
\newblock {\em arXiv}, 2019.

\bibitem{DBLP:journals/corr/KingmaB14}
Diederik~P. Kingma and Jimmy Ba.
\newblock Adam: {A} method for stochastic optimization.
\newblock {\em CoRR}, abs/1412.6980, 2014.

\bibitem{kingma2013auto}
Diederik~P Kingma and Max Welling.
\newblock Auto-encoding variational bayes.
\newblock {\em arXiv preprint arXiv:1312.6114}, 2013.

\bibitem{Li:2018:LDGG}
Yujia Li, Oriol Vinyals, Chris Dyer, Razvan Pascanu, and Peter Battaglia.
\newblock Learning deep generative models of graphs.
\newblock In {\em ICLR}, 2018.

\bibitem{GraphAttention2019}
Renjie Liao, Yujia Li, Yang Song, Shenlong Wang, Will Hamilton, David~K
  Duvenaud, Raquel Urtasun, and Richard Zemel.
\newblock Efficient graph generation with graph recurrent attention networks.
\newblock In H. Wallach, H. Larochelle, A. Beygelzimer, F. d\textquotesingle
  Alch\'{e}-Buc, E. Fox, and R. Garnett, editors, {\em Advances in Neural
  Information Processing Systems}, volume~32, pages 4255--4265. Curran
  Associates, Inc., 2019.

\bibitem{Liu:2017:RasterToVector}
Chen Liu, Jiajun Wu, Pushmeet Kohli, and Yasutaka Furukawa.
\newblock Raster-to-vector: Revisiting floorplan transformation.
\newblock {\em ICCV}, 2017.

\bibitem{GraphFlows2019}
Jenny Liu, Aviral Kumar, Jimmy Ba, Jamie Kiros, and Kevin Swersky.
\newblock Graph normalizing flows.
\newblock In H. Wallach, H. Larochelle, A. Beygelzimer, F. d\textquotesingle
  Alch\'{e}-Buc, E. Fox, and R. Garnett, editors, {\em Advances in Neural
  Information Processing Systems}, volume~32, pages 13578--13588. Curran
  Associates, Inc., 2019.

\bibitem{Majerowicz:2014:FYS}
L. Majerowicz, A. Shamir, A. Sheffer, and H.~H. Hoos.
\newblock Filling your shelves: Synthesizing diverse style-preserving artifact
  arrangements.
\newblock {\em IEEE Transactions on Visualization and Computer Graphics}, 2014.

\bibitem{bookStreet}
S. Marshall.
\newblock {\em Streets and Patterns}.
\newblock Routledge, 2015.

\bibitem{Merrell:2010:CGR}
Paul Merrell, Eric Schkufza, and Vladlen Koltun.
\newblock Computer-generated residential building layouts.
\newblock {\em ACM Trans. Graph.}, 29(6):181:1--181:12, July 2010.

\bibitem{Merrell:2011:IFL}
Paul Merrell, Eric Schkufza, Zeyang Li, Maneesh Agrawala, and Vladlen Koltun.
\newblock Interactive furniture layout using interior design guidelines.
\newblock {\em ACM Trans. Graph.}, 30(4):87:1--87:10, July 2011.

\bibitem{Mueller:2006:PMO}
Pascal M{\"u}ller, Peter Wonka, Simon Haegler, Andreas Ulmer, and Luc Van~Gool.
\newblock Procedural modeling of buildings.
\newblock {\em ACM Transactions on Graphics}, 25(3):614--623, 2006.

\bibitem{nash2020polygen}
Charlie Nash, Yaroslav Ganin, SM Eslami, and Peter~W Battaglia.
\newblock Polygen: An autoregressive generative model of 3d meshes.
\newblock {\em arXiv preprint arXiv:2002.10880}, 2020.

\bibitem{housegan2020}
Nelson Nauata, Kai-Hung Chang, Chin-Yi Cheng, Greg Mori, and Yasutaka Furukawa.
\newblock House-gan: Relational generative adversarial networks for
  graph-constrained house layout generation.
\newblock 2020.

\bibitem{lifull}
National~Institute of Informatics.
\newblock {LIFULL HOME'S Dataset}, 2020.

\bibitem{GraphEmbedding2020}
S. {Pan}, R. {Hu}, S. {Fung}, G. {Long}, J. {Jiang}, and C. {Zhang}.
\newblock Learning graph embedding with adversarial training methods.
\newblock {\em IEEE Transactions on Cybernetics}, 50(6):2475--2487, 2020.

\bibitem{parmar2018image}
Niki Parmar, Ashish Vaswani, Jakob Uszkoreit, {\L}ukasz Kaiser, Noam Shazeer,
  Alexander Ku, and Dustin Tran.
\newblock Image transformer.
\newblock {\em arXiv preprint arXiv:1802.05751}, 2018.

\bibitem{paszke2017automatic}
Adam Paszke, Sam Gross, Soumith Chintala, Gregory Chanan, Edward Yang, Zachary
  DeVito, Zeming Lin, Alban Desmaison, Luca Antiga, and Adam Lerer.
\newblock Automatic differentiation in pytorch.
\newblock 2017.

\bibitem{Peng:2016:CND}
Chi-Han Peng, Yong-Liang Yang, Fan Bao, Daniel Fink, Dong-Ming Yan, Peter
  Wonka, and Niloy~J Mitra.
\newblock Computational network design from functional specifications.
\newblock {\em ACM Transactions on Graphics}, 35(4):131, 2016.

\bibitem{Peng:2014:CLW}
Chi-Han Peng, Yong-Liang Yang, and Peter Wonka.
\newblock Computing layouts with deformable templates.
\newblock {\em ACM Transactions on Graphics}, 33(4):99, 2014.

\bibitem{Prusinkiewicz:1990:ABP}
Przemyslaw Prusinkiewicz and Aristid Lindenmayer.
\newblock {\em The Algorithmic Beauty of Plants}.
\newblock Springer-Verlag, New York, 1990.

\bibitem{radford2019language}
Alec Radford, Jeff Wu, Rewon Child, David Luan, Dario Amodei, and Ilya
  Sutskever.
\newblock Language models are unsupervised multitask learners.
\newblock 2019.

\bibitem{DBLP:conf/nips/RazaviOV19}
Ali Razavi, A{\"{a}}ron van~den Oord, and Oriol Vinyals.
\newblock Generating diverse high-fidelity images with {VQ-VAE-2}.
\newblock In {\em Advances in Neural Information Processing Systems 32: Annual
  Conference on Neural Information Processing Systems 2019, NeurIPS 2019, 8-14
  December 2019, Vancouver, BC, Canada}, pages 14837--14847, 2019.

\bibitem{rezende2015variational}
Danilo~Jimenez Rezende and Shakir Mohamed.
\newblock Variational inference with normalizing flows.
\newblock {\em arXiv preprint arXiv:1505.05770}, 2015.

\bibitem{richardson2020encoding}
Elad Richardson, Yuval Alaluf, Or Patashnik, Yotam Nitzan, Yaniv Azar, Stav
  Shapiro, and Daniel Cohen-Or.
\newblock Encoding in style: a stylegan encoder for image-to-image translation.
\newblock {\em arXiv preprint arXiv:2008.00951}, 2020.

\bibitem{Rubner:1998:MDA}
Yossi Rubner, Carlo Tomasi, and Leonidas~J. Guibas.
\newblock A metric for distributions with applications to image databases.
\newblock In {\em Proc. {ICCV}}, ICCV '98, 1998.

\bibitem{Simonovsky:2018:GraphVAE}
Martin Simonovsky and Nikos Komodakis.
\newblock {GraphVAE:} towards generation of small graphs using variational
  autoencoders.
\newblock In {\em ICLR}, 2018.

\bibitem{JMLR:v15:srivastava14a}
Nitish Srivastava, Geoffrey Hinton, Alex Krizhevsky, Ilya Sutskever, and Ruslan
  Salakhutdinov.
\newblock Dropout: A simple way to prevent neural networks from overfitting.
\newblock {\em Journal of Machine Learning Research}, 15(56):1929--1958, 2014.

\bibitem{Tarabishy2019DeepLS}
Sherif Tarabishy, Stamatios Psarras, Marcin Kosicki, and Martha Tsigkari.
\newblock Deep learning surrogate models for spatial and visual connectivity.
\newblock {\em ArXiv}, 2019.

\bibitem{vahdat2020NVAE}
Arash Vahdat and Jan Kautz.
\newblock {NVAE}: A deep hierarchical variational autoencoder.
\newblock In {\em Advances in Neural Information Processing Systems}, 2020.

\bibitem{Vanegas:2012:PGO}
Carlos~A. Vanegas, Tom Kelly, Basil Weber, Jan Halatsch, Daniel Aliaga, and
  Pascal M{\"u}ller.
\newblock Procedural generation of parcels in urban modeling.
\newblock {\em Computer Graphics Forum}, 31(2), 2012.

\bibitem{vaswani2017transformers}
Ashish Vaswani, Noam Shazeer, Niki Parmar, Jakob Uszkoreit, Llion Jones,
  Aidan~N Gomez, {\L}ukasz Kaiser, and Illia Polosukhin.
\newblock Attention is all you need.
\newblock In {\em Advances in Neural Information Processing Systems}, pages
  5998--6008, 2017.

\bibitem{vinyals2015pointer}
Oriol Vinyals, Meire Fortunato, and Navdeep Jaitly.
\newblock Pointer networks.
\newblock In {\em Advances in Neural Information Processing Systems}, pages
  2692--2700, 2015.

\bibitem{Wang:2018:AAAI:graphGAN}
Hongwei Wang, Jia Wang, Jialin Wang, Miao Zhao, Weinan Zhang, Fuzheng Zhang,
  Xing Xie, and Minyi Guo.
\newblock Graphgan: Graph representation learning with generative adversarial
  nets.
\newblock In {\em AAAI}, 2018.

\bibitem{PlanIT:2019:Wang}
Kai Wang, Yu{-}An Lin, Ben Weissmann, Manolis Savva, Angel~X. Chang, and Daniel
  Ritchie.
\newblock Planit: planning and instantiating indoor scenes with relation graph
  and spatial prior networks.
\newblock {\em ACM Transactions on Graphics}, 2019.

\bibitem{Wang:2018:DCP}
Kai Wang, Manolis Savva, Angel~X. Chang, and Daniel Ritchie.
\newblock Deep convolutional priors for indoor scene synthesis.
\newblock {\em ACM Trans. Graph.}, 37(4):70:1--70:14, July 2018.

\bibitem{Wu:2018:MLD}
Wenming Wu, Lubin Fan, Ligang Liu, and Peter Wonka.
\newblock Miqp-based layout design for building interiors.
\newblock {\em Computer Graphics Forum}, 2018.

\bibitem{Wu_DeepLayout_2019}
Wenming Wu, Xiao-Ming Fu, Rui Tang, Yuhan Wang, Yu-Hao Qi, and Ligang Liu.
\newblock Data-driven interior plan generation for residential buildings.
\newblock {\em ACM Transactions on Graphics}, 2019.

\bibitem{layernorm19}
Jingjing Xu, Xu Sun, Zhiyuan Zhang, Guangxiang Zhao, and Junyang Lin.
\newblock Understanding and improving layer normalization.
\newblock In H. Wallach, H. Larochelle, A. Beygelzimer, F. d\textquotesingle
  Alch\'{e}-Buc, E. Fox, and R. Garnett, editors, {\em Advances in Neural
  Information Processing Systems}, volume~32, pages 4381--4391. Curran
  Associates, Inc., 2019.

\bibitem{yang2019pointflow}
Guandao Yang, Xun Huang, Zekun Hao, Ming-Yu Liu, Serge Belongie, and Bharath
  Hariharan.
\newblock Pointflow: 3d point cloud generation with continuous normalizing
  flows.
\newblock In {\em Proceedings of the IEEE International Conference on Computer
  Vision}, pages 4541--4550, 2019.

\bibitem{Yang:2013:UPL}
Yong-Liang Yang, Jun Wang, Etienne Vouga, and Peter Wonka.
\newblock Urban pattern: Layout design by hierarchical domain splitting.
\newblock {\em ACM Transactions on Graphics}, 32(6), 2013.

\bibitem{Yeh:2013:STP}
Yi-Ting Yeh, Katherine Breeden, Lingfeng Yang, Matthew Fisher, and Pat
  Hanrahan.
\newblock Synthesis of tiled patterns using factor graphs.
\newblock {\em ACM Trans. Graph.}, 32(1):3:1--3:13, Feb. 2013.

\bibitem{Yeh:2012:SOW}
Yi-Ting Yeh, Lingfeng Yang, Matthew Watson, Noah~D. Goodman, and Pat Hanrahan.
\newblock Synthesizing open worlds with constraints using locally annealed
  reversible jump mcmc.
\newblock {\em ACM Transactions on Graphics}, 31(4), 2012.

\bibitem{You:2018:GraphRNN}
Jiaxuan You, Rex Ying, Xiang Ren, William Hamilton, and Jure Leskovec.
\newblock {GraphRNN:} generating realistic graphs with deep auto-regressive
  models.
\newblock In {\em ICML}, pages 5694--5703, 2018.

\bibitem{Yu:2011:MakeItHome}
Lap-Fai Yu, Sai-Kit Yeung, Chi-Keung Tang, Demetri Terzopoulos, Tony~F. Chan,
  and Stanley~J. Osher.
\newblock Make it home: Automatic optimization of furniture arrangement.
\newblock {\em ACM Trans. Graph.}, 30(4):86:1--86:12, July 2011.

\bibitem{Yu:2011:MHA}
Lap-Fai Yu, Sai-Kit Yeung, Chi-Keung Tang, Demetri Terzopoulos, Tony~F. Chan,
  and Stanley~J. Osher.
\newblock Make it home: Automatic optimization of furniture arrangement.
\newblock {\em ACM Trans. Graph.}, 30(4):86:1--86:12, July 2011.

\bibitem{Han18SAG}
Han Zhang, Ian~J. Goodfellow, Dimitris~N. Metaxas, and Augustus Odena.
\newblock Self-attention generative adversarial networks.
\newblock {\em arXiv:1805.08318}, 2018.

\bibitem{Zhu2017CycleGAN}
Jun-Yan Zhu, Taesung Park, Phillip Isola, and Alexei~A. Efros.
\newblock Unpaired image-to-image translation using cycle-consistent
  adversarial networks.
\newblock In {\em 2017 IEEE International Conference on Computer Vision
  (ICCV)}, 2017.

\bibitem{NIPS2017_6650}
Jun-Yan Zhu, Richard Zhang, Deepak Pathak, Trevor Darrell, Alexei~A Efros,
  Oliver Wang, and Eli Shechtman.
\newblock Toward multimodal image-to-image translation.
\newblock In I. Guyon, U.~V. Luxburg, S. Bengio, H. Wallach, R. Fergus, S.
  Vishwanathan, and R. Garnett, editors, {\em Advances in Neural Information
  Processing Systems 30}, pages 465--476. Curran Associates, Inc., 2017.

\bibitem{zhu2019sean}
Peihao Zhu, Rameen Abdal, Yipeng Qin, and Peter Wonka.
\newblock Sean: Image synthesis with semantic region-adaptive normalization.
\newblock 2020.

\end{thebibliography}
}

\appendix

\section{Furniture Layout Implementation Details}
\label{sec:furniture_details}
In this Section, we describe the implementation details for furniture layouts that differ from floor plans. Since furniture layouts are less constrained than floor plans (furniture pieces do not need to cover all of the layout without gaps, for example), we do not add constraining edges and omit the optimization step, directly using the element constraints as elements instead: $N = N^C$.

\paragraph{Layout representation}
In furniture layouts, each element represents a piece of furniture with an oriented bounding box $N=(\tau, x, y, w, h, \alpha)$ that is parameterized by the type of furniture $\tau$, the lower-left corner of the bounding box $(x,y)$, the width and height of the bounding box $(w,h)$, and its orientation $\alpha$.

\paragraph{Element constraints}
The element constraint model described in Section 3.2 of the main paper generates constraints $N^C = (\tau, x, y, w, h, \alpha)$ for all parameters of a furniture piece that are directly used as furniture pieces $N$.


\begin{table}[t]
\caption{Free generation of furniture layouts. We compare the layout statistics of furniture layouts generated by our method to purely image-based generation with StyleGAN \cite{styleGan18}. Our method shows a clear improvement over image-based generation.}
\centering 
\setlength{\tabcolsep}{5pt}
\begin{tabular}{@{}lS[table-format=2.2]S[table-format=2.2]S[table-format=2.2]S[table-format=2.2]c@{}} 
\toprule
method & $\hat{s}_{t}$ & $\hat{s}_{r}$ & $\hat{s}_{a}$ & $\hat{\mathbf{s}}_\textbf{avg}$ \\
\midrule \midrule
StyleGAN & 16.50 & 6.24 & 7.09 & 9.94 & \\
ours free & \boldentry{1.00} & \boldentry{1.00} & \boldentry{1.00} & \boldentry{1.00}
 \\
\bottomrule
\end{tabular}
\label{tab:furniture_free_gen}
\end{table}

\begin{figure*}[t]
    \centering
    \includegraphics[width=\textwidth]{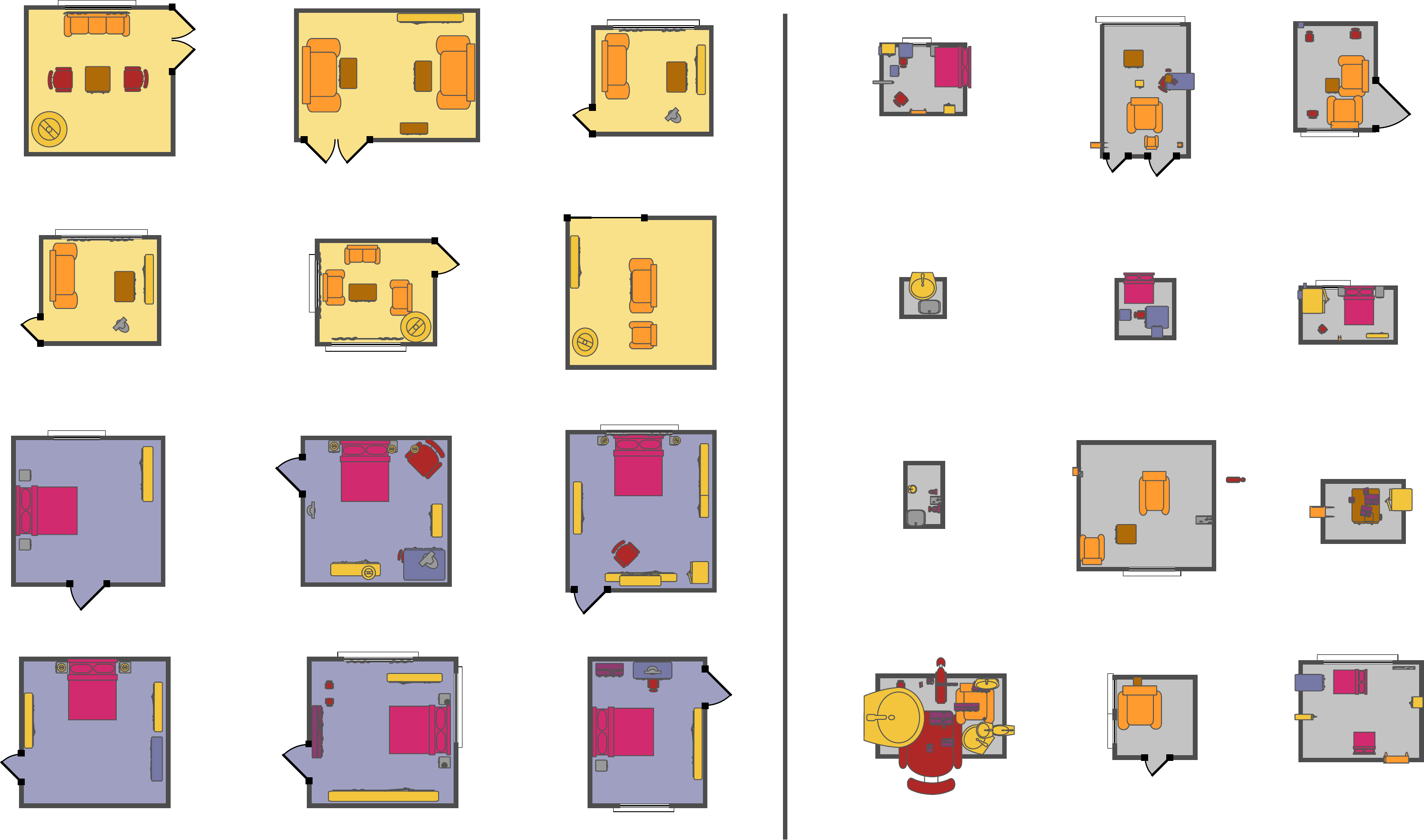}
    \caption{Additional furniture layout results compared to StyleGAN. Left: our furniture layouts (yellow: living-room; blue: bedroom); right: StyleGAN does not generate correct furniture proportions and has a lot of noise in its layouts.}
    \label{fig:furniture_free_gen}
\end{figure*}

\begin{figure*}[t]
    \centering
    \includegraphics[height=3.6in]{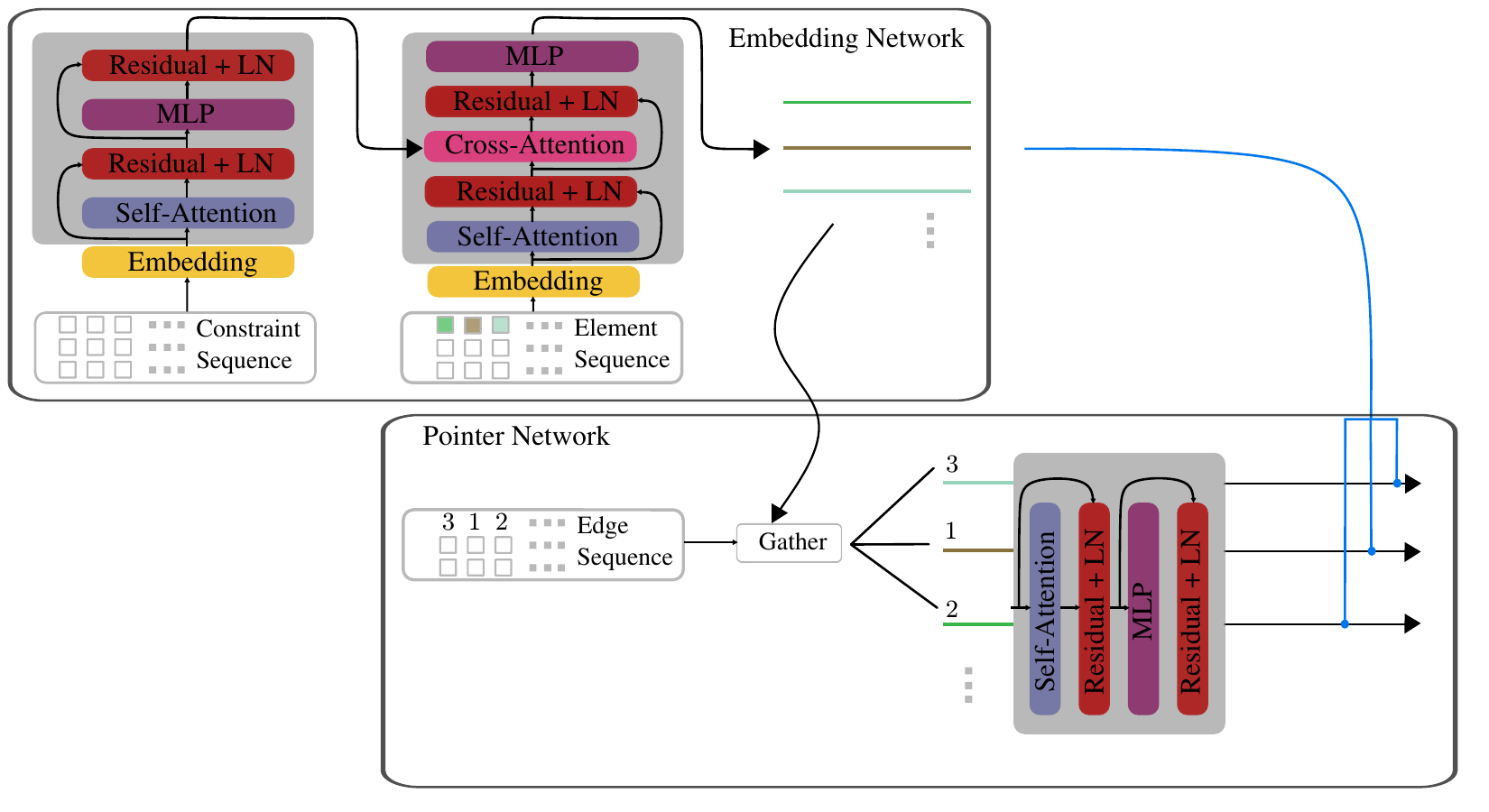}
    \caption{The user-constrained Edge Generation Model.
    An embedding function modeled by a transformer (top left) generates element embeddings that are re-arranged based on the edge sequence. This sequence is ingested by the edge model (bottom right), which is also implemented as a transformer. The encoder (left block in the embedding network) is only used when performing constrained generation.
    }
    \label{fig:edge_models}
\end{figure*}

\section{Additional Furniture Layout Results}
\label{sec:layout_results}
In this section, we present additional furniture layout results. We generated approximately 10k furniture layouts for all room types in our floor plans. We evaluate these furniture layouts using the layout statistics described in Section 4 of the main paper. To compute topological statistics $\hat{s}_t$, we create an $r$-NN graph of the furniture pieces as layout graph, with $r=15\%$ of the layout diagonal. Thus, topological statistics capture relationships in local neighborhoods of furniture pieces, for example which types of furniture are typically placed next to each other. Since elements in furniture layouts have additional parameters, we extend the list of layout statistics. We add one statistic to the shape statistics $S_r$:
\begin{itemize}[itemsep=-3pt]
    \small
    \item[$s_r^o$:] a histogram of orientation distributions for each element type.
\end{itemize}
And the alignment statistics $S_a$ are extended with:
\begin{itemize}[itemsep=-3pt]
    \small
    \item[$s_s^o$:] a histogram of the differences between orientations.
    \item[$s_s^w$:] a histogram of the differences between widths.
    \item[$s_s^h$:] a histogram of the differences between heights.
\end{itemize}

We compare to furniture layouts generated with StyleGAN. Similar to floor plans, we render our furniture layout dataset, train StyleGAN, and parse the generated images back into furniture layouts. Table~\ref{tab:furniture_free_gen} and Figure~\ref{fig:furniture_free_gen} show the results of this comparison. Similar to floor plans, Our method shows a clear advantage over the purely image-based StyleGAN.




\section{Architecture Details}
\label{sec:architecture_details}
In this section, we describe the units we use as the building blocks for our model. 

\subsection{Embedding}
\paragraph{Element Constraint Model} The input sequence to the element constraint model has three components - the value sequence $S_E = \{v_i\}_{i=1}^{kM_N}$, the position sequence $I = \{i\}_{i=1}^{kM_N}$ and the type sequence $T = \{i\, \text{mod}\, k \}_{i=1}^{kM_N}$, where $M_N$ is the number of elements and $k$ is the number of properties per element. We use three separate learned embeddings, one for each sequence. The final embedding is the sum of these three embeddings.

\paragraph{Edge Model} The edge model operates on sequences of learned element embeddings $g_{\theta_\rho}$, as described in Section 3.3 of the paper. The embedding function is modeled by a transformer with the same architecture as the element constraint model, that takes as input the element constraint sequence and outputs a sequence of element embeddings. Similar to the element constraint model, the embedding function  can be conditioned on a sequence of constraints by adding an encoder, as shown in the top left of Figure~\ref{fig:edge_models}.

The sequence of element embeddings is then arranged according to the edge sequence (concatenating the element embeddings corresponding to the two elements of each edge) and processed by the edge model (Figure~\ref{fig:edge_models}, right) as described in Section 3.3 of the paper.

\subsection{GPT2- Blocks}

For completeness, we describe the details of the architecture given in Figure 4 of the main paper.
The yellow embedding block denotes the embedding of the element constraint model, as described above.
We use Dropout \cite{JMLR:v15:srivastava14a} with a drop probability of $0.2$ immediately after performing the sum of embeddings. The attention layers in all our experiments use Multiheaded Attention with 12 heads. We set our embedding dimension $d = 384$.


\paragraph{Encoder}
We use a stack of standard GPT-2 \cite{radford2019language} encoder blocks.
The MLP block inside the encoder (and the decoder) performs the following operation on an input tensor $x$
\begin{equation}
    x = \text{Linear}(\text{GELU} (\text{Linear}(x)))
\end{equation}

The activation function we use between the linear layers is the GELU \cite{hendrycks2016gelu} function. The first linear layer changes the embedding dimensions internally from $d$ to $4d$. The second then goes back from $4d$ to $d$



\paragraph{Decoder}
The activation $h^{(L)}$ obtained at the last layer of the encoder is used for performing cross-attention in the Decoder.
We can write the operations of a Decoder block as:
\begin{equation}
    {n}^{(i)} = \text{LN} (h_D^{(i)})
\end{equation}
\begin{equation}
    {a}^{(i)} = \text{LN}\left( {n}^{(i)} + \text{SelfAttention}(n^{(i)}, n^{(i)})\right)
\end{equation}
\begin{equation}
    {b}^{(i)} = \text{LN}\left( {a}^{(i)} + \text{CrossAttention}(a^{(i)}, h^{(L)})\right)
\end{equation}
\begin{equation}
    {h_D}^{(i+1)} = b^{(i)} + \text{MLP}(b^{(i)}),
\end{equation}
where LN denotes Layer Normalization~\cite{ba2016layer}.
We a add a single linear layer after both the Encoder and the Decoder to produce logits.
The encoders are only used for constrained generation, such as floor plan generation constrained on a given floor plan boundary. In free generation, we do not have any constraints, so we do not add encoders to any of the models.




\end{document}